\documentclass[journal]{IEEEtran}

\usepackage{graphicx}
\usepackage[toc]{glossaries}

\usepackage{float}
\restylefloat{table}

\usepackage[hidelinks]{hyperref}
\hypersetup{
    colorlinks=true,
    linkcolor=blue,
    filecolor=magenta,
    urlcolor=cyan,
}

\usepackage{amsmath}
\usepackage{amssymb}
\usepackage{eucal}
\usepackage{booktabs}
\usepackage{caption,subcaption}
\usepackage{changes}
\usepackage{pifont}
\usepackage{rotating}
\usepackage{framed}

\usepackage{type1cm}
\usepackage{lettrine}

\usepackage{tikz}
\usetikzlibrary{backgrounds}

\let\labelindent\relax
\usepackage{enumitem}

\DeclareMathAlphabet\mathbfcal{OMS}{cmsy}{b}{n}

\usepackage{tcolorbox}
\usepackage{nicefrac,xfrac}

\DeclareMathOperator*{\relu}{\textsc{ReLU}}

\DeclareMathOperator*{\argmaxA}{arg\,max} 

\newcommand{\figref}[1]{Figure~\ref{#1}}
\newcommand{\tableref}[1]{Table~\ref{#1}}
\renewcommand{\eqref}[1]{Equation~\ref{#1}}
\newcommand{\secref}[1]{Section~\ref{#1}}

\newcommand{\mycomment}[1]{}

\usepackage{adjustbox}
\usepackage{array}
\usepackage{multirow}
\newcolumntype{L}[1]{>{\raggedright\let\newline\\\arraybackslash\hspace{0pt}}m{#1}}
\newcolumntype{C}[1]{>{\centering\let\newline\\\arraybackslash\hspace{0pt}}m{#1}}

\newcommand{\MV}[1]{{\color{blue}{}}}
\newcommand{\cmt}[1]{{\color{cyan}{}}}

\newcolumntype{R}[2]{%
    >{\adjustbox{angle=#1,lap=\width-(#2)}\bgroup}%
    c%
    <{\egroup}%
}
\newcommand*\rot{\multicolumn{1}{R{30}{1em}}}
\newcommand*\rotso{\multicolumn{1}{R{45}{1em}}}

\makeglossaries

\newacronym{drs}{DRS}{Diabetic Retinopathy Screening}
\newacronym{vqa}{VQA}{Visual Question Answering}
\newacronym{slam}{SLAM}{Simultaneous Localization and Mapping}
\newacronym[plural=CNNs,firstplural=Convolutional Neural Networks (CNNs)]{cnn}{CNN}{Convolutional Neural Network}
\newacronym{nlp}{NLP}{Natural-language processing}
\newacronym{mcb}{MCB}{Multimodal Compact Bilinear}
\newacronym{mlb}{MLB}{Multimodal Low-rank Bilinear}
\newacronym{mutan}{MUTAN}{Multimodal Tucker Fusion for Visual Question Answering}
\newacronym{idrid}{IDRID}{Indian Diabetic Retinopathy Image Dataset}
\newacronym[plural=RNNs,firstplural=Recurrent Neural Networks (RNNs)]{rnn}{RNN}{Recurrent Neural Network}
\newacronym{lstm}{LSTM}{Long Short-Term Memory}
\newacronym{bow}{BOW}{bag-of-words}
\newacronym{gru}{GRU}{Gated Recurrent Units}
\newacronym[plural=QAs,firstplural=Questions \& Answers (QAs)]{qa}{QA}{Question \& Answer}
\newacronym{ma}{MA}{Microaneurysms}
\newacronym{he}{HE}{Hemorrhages}
\newacronym{ex}{EX}{Hard Exudates}
\newacronym{se}{SE}{Soft Exudates}
\newacronym{nar}{NAR}{no apparent retinopathy}
\newacronym{dr}{DR}{Diabetic Retinopathy}
\newacronym{dme}{DME}{Diabetic Macular Edema}
\newacronym{pdr}{PDR}{Proliferative Diabetic Retinopathy}
\newacronym{wmlb}{WMLB}{Weighted Multimodal Low-rank Bilinear Attention Network}
\newacronym{relu}{ReLU}{Rectified Linear Unit}
\newacronym{dl}{DL}{deep learning}
\newacronym{qcmlb}{QC-MLB}{Question-Centric Multimodal Low-rank Bilinear}
\newacronym{bleu}{BLEU}{Bilingual Evaluation Understudy}
\newacronym{gmlb}{G-MLB}{Global Multimodal Low-rank Bilinear}
\newacronym{bert}{BERT}{Bidirectional Encoder Representations from Transformers}
\newacronym{gradcam}{Grad-CAM}{Gradient-weighted Class Activation Mapping}
\newacronym{xlnet}{XLNet}{Generalized Autoregressive Pretraining for Language Understanding}
\newacronym[plural=GPUs,firstplural=graphics processing units (GPUs)]{gpu}{GPU}{graphics processing units}

\usepackage{scalefnt}

\definecolor{wrongColor}{RGB}{229,57,53}
\definecolor{rightColor}{RGB}{56,142,60}

\begin{document}
%


\title{A Question-Centric Model for Visual Question Answering in Medical Imaging}

\author{Minh H. Vu,
        Tommy L\"{o}fstedt,
        Tufve Nyholm,
        and Raphael Sznitman
\thanks{Copyright (c) 2019 IEEE. Personal use of this material is permitted. However, permission to use this material for any other purposes must be obtained from the IEEE by sending a request to pubs-permissions@ieee.org.}        
\thanks{This research was conducted using the resources of the High Performance Computing Center North (HPC2N) at Ume{\aa} University, Ume{\aa}, Sweden. We are grateful for the financial support obtained from the Cancer Research Fund in Northern Sweden, Karin and Krister Olsson, Ume\aa{} University, The V\"{a}sterbotten regional county, and Vinnova, the Swedish innovation agency.}
\thanks{$^{1}$M. Vu, T. L\"{o}fstedt, and T. Nyholm are with the Department of Radiation Sciences, Ume{\aa} University, Ume{\aa}, Sweden. E-mail: minh.vu@umu.se.}
\thanks{$^{2}$R. Sznitman is with the ARTORG Center, University of Bern, Switzerland.}
}

\maketitle              

\begin{abstract}
Deep learning methods have proven extremely effective at performing a variety of medical image analysis tasks. With their potential use in clinical routine, their lack of transparency has however been one of their few weak points, raising concerns regarding their behavior and failure modes. While most research to infer model behavior has focused on indirect strategies that estimate prediction uncertainties and visualize model support in the input image space, the ability to explicitly query a prediction model regarding its image content offers a more direct way to determine the behavior of trained models. To this end, we present a novel Visual Question Answering approach that allows an image to be queried by means of a written question. Experiments on a variety of medical and natural image datasets show that by fusing image and question features in a novel way, the proposed approach achieves an equal or higher accuracy compared to current methods.
\end{abstract}


\section{Introduction}
\label{sec:intro}

\lettrine{D}{eep} learning has fundamentally reshaped medical image analysis with methods that perform at remarkable levels. With excellent results for numerous diagnostic tasks, this advancement has caught the attention of clinical communities such as radiology, pathology, and ophthalmology. At the same time, the black-box nature of \gls{dl} methods has also raised many concerns as to what these methods are doing, what their biases are, and when and where they ultimately fail.

To alleviate these concerns, considerable research has focused on providing better answers to how 
\gls{dl} methods behave under different conditions. This has primarily consisted of methods capable of providing prediction uncertainties or visualization of evidence in the input image space. For instance, Bayesian dropout or ensembles provide computational approaches to assess prediction uncertainty from trained models. Alternatively, saliency or \gls{gradcam}~\cite{Selvaraju_etal_2017} maps are now standard ways to gain insight into what part of the input data is used by a \gls{cnn} for a particular prediction. These methods thus allow indirect interpretations of \gls{dl} models.

\begin{figure*}[t!]

    \def\subfigsize{0.32\textwidth}
    \def\subfigverlabelsize{0.16\textwidth}
    \def\scalefonesize{1}
    \def\scalefonesizetitle{1.1}

    \centering

    \captionsetup[subfigure]{labelformat=empty, justification=centering}
    \begin{subfigure}{\subfigsize}
        \caption{{\scalefont{\scalefonesizetitle} Are there soft exudates in the fundus?}}
    \end{subfigure} 
    \begin{subfigure}{\subfigsize}
        \caption{{\scalefont{\scalefonesizetitle}What is the major class in the image?}}
    \end{subfigure} 
    \begin{subfigure}{\subfigsize}
        \caption{{\scalefont{\scalefonesizetitle}How many tools are there?}}
    \end{subfigure} \\
    
    \captionsetup[subfigure]{labelformat=empty, justification=centering}
    \begin{subfigure}{\subfigverlabelsize}
        \caption{{\scalefont{\scalefonesizetitle}no}}
    \end{subfigure} 
    \begin{subfigure}{\subfigverlabelsize}
        \caption{{\scalefont{\scalefonesizetitle}yes}}
    \end{subfigure} 
    \begin{subfigure}{\subfigverlabelsize}
        \caption{{\scalefont{\scalefonesizetitle}invasive}}
    \end{subfigure}
    \captionsetup[subfigure]{labelformat=empty, justification=centering}
    \begin{subfigure}{\subfigverlabelsize}
        \caption{{\scalefont{\scalefonesizetitle}normal}}
    \end{subfigure} 
    \begin{subfigure}{\subfigverlabelsize}
        \caption{{\scalefont{\scalefonesizetitle}3}}
    \end{subfigure} 
    \begin{subfigure}{\subfigverlabelsize}
        \caption{{\scalefont{\scalefonesizetitle}2}}
    \end{subfigure} \\    
    
    \includegraphics[width=\subfigverlabelsize]{figures/vqa1.jpg}
    \includegraphics[width=\subfigverlabelsize]{figures/vqa2.jpg}
    \includegraphics[width=\subfigverlabelsize]{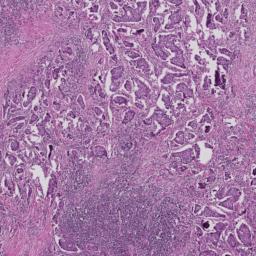}
    \includegraphics[width=\subfigverlabelsize]{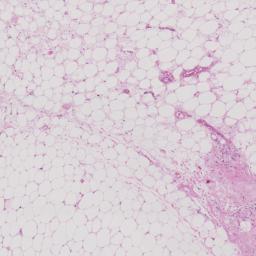}
    \includegraphics[width=\subfigverlabelsize]{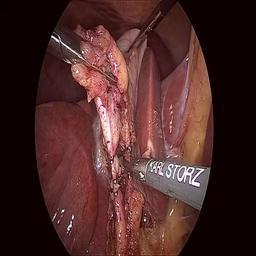}
    \includegraphics[width=\subfigverlabelsize]{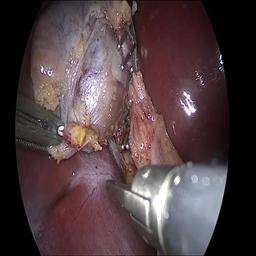}    
    \\

    \caption{Examples of questions and their corresponding answers in three medical contexts: Retinopathy Screening (left), Breast Cancer Grading (center), and Surgical Tools Detection (right).}
    \label{fig:example}
\end{figure*}

A recent alternative is the \gls{vqa} concept~\cite{kafle2017visual,ImageCLEF19}, whereby a model is queried regarding the content of an image by means of an explicit question. Here, a neural network (NN) model processes both the image and the question directly and outputs an answer to the question. As such, it allows direct model probing with respect to the inputs, in much the same way as a {\it Turing Test}~\cite{geman2015visual,fountoukidou2019}. \gls{vqa} thus allows a trained model to be explicitly tested regarding a variety of concepts and content to characterize its behavior (see \figref{fig:example}).

In this context, we present a novel \gls{vqa} approach applied to medical image data. As inputs, our framework receives an image and a question in the form of a written sentence, and outputs an answer to the question in one of the following formats: binary (\textit{e.g.}, ``yes'' or ``no''), numbers, categories, locations, or short sentences. The approach proposed here, denoted \gls{qcmlb}, fuses image and question features by enforcing high adherence to the query sentence. The proposed \gls{vqa} approach is shown to perform at worst equally well as state-of-the-art methods on four medical imaging datasets, and two natural image datasets. Furthermore, we demonstrate that the proposed \gls{vqa} approach is easily combined with CAM-like methods to highlight which parts of the image are used by the model to answer the question, further aiding the understanding of how the trained model works.

In the present work, our main contributions are: (i) we introduce a \textit{question-centric} multimodal fusion scheme between image and question representations, that emphasizes the question features, for the \gls{vqa} task; (ii) we present \gls{vqa} applications for medical imaging; and (iii) we propose an approach to generate \gls{vqa} ground truth pairs for specific medical areas, that can be extended to any other field.

\section{Related Works}
\label{sec:related}

Recent works exploring \gls{vqa} methods in the medical domain have mainly come from contributions to the recent Visual Question Answering challenge in the Medical Domain hosted by ImageCLEF~\cite{ImageCLEFVQA-Med2019}. Overall, however, state-of-the-art methods used for medical imaging have paralleled those in natural images very closely. In particular, different \gls{vqa} models have focused on how they integrate the question and image inputs in the model. Various \gls{vqa} techniques were reviewed in~\cite{kafle2017visual}, where the recent approaches were found to be,
\begin{itemize}[]
    \item using Bayesian models to exploit the underlying relationships between question-image-answer feature distributions~\cite{shih2016look},
    \item using the question to break the VQA task into a sequence of modular sub-problems~\cite{andreas2016learning}. For example, the question ``what is the major class in the image?'' (see \figref{fig:example}) maybe answered in two consecutive steps: ``how many classes are there?'' and ``which class outnumbers the rest?'',
    \item combining the image and question features using pooling schemes in a neural network framework~\cite{fukui2016multimodal,kim2016hadamard,ben2017mutan,minh2019vqa}.
\end{itemize}

Most \gls{vqa} models using pooling~\cite{kim2016hadamard} in a neural network framework include five core learnable elements: (i) a question model with the purpose of encoding the question inputs, (ii) an image model that extracts visual features from the input images, (iii) a fusion scheme that combines the visual and question features, (iv) an attention scheme that looks for important regions in the input image, and (v) a classifier that selects the top answer among a set of candidates. 

\subsubsection{Question Model} \gls{lstm}~\cite{hochreiter1997long}, \gls{gru}~\cite{cho2014learning}, and Skip-thought vectors~\cite{kiros2015skip} are commonly employed to extract question features. Skip-thought vectors are an unsupervised encoder-decoder approach that learns the semantics and continuity of text from novels to reconstruct the surrounding sentences of a given encoded text. Eventually, sentences, which are semantic-related, are represented by a similar vector. One of the key advantages of the Skip-thought model, and possibly one of the main reasons for its success, is the ability to expand its vocabulary. Specifically, the pre-trained model of Skip-thought vectors can be `fine-tuned' to learn a rich vocabulary (up to millions of words).

Skip-thought vectors are a powerful approach that has been used in many recent \gls{vqa} models~\cite{fukui2016multimodal,kim2016hadamard,ben2017mutan}. In the present work, we also use Skip-thought vectors in order to extract word features. Recently, Google AI\footnote{\url{https://ai.google/research/}} presented a pre-trained \gls{bert} model~\cite{devlin2018bert} and a \gls{xlnet}~\cite{yang2019xlnet} that achieved state-of-the-art results on 18 \gls{nlp} tasks.

\subsubsection{Image Model} The image model is used to extract visual features from the input images. Most recent \gls{vqa} systems use \glspl{cnn}, that are pre-trained (\textit{e.g.}\ the ImageNet dataset~\cite{krizhevsky2012imagenet}). Common choices for the image model comprise: VGGNet~\cite{simonyan2014very}, GoogLeNet~\cite{szegedy2015going}, and ResNet~\cite{fukui2016multimodal,kim2016hadamard,ben2017mutan}. Similar to \gls{mcb}~\cite{fukui2016multimodal}, \gls{mlb}~\cite{kim2016hadamard}, \gls{mutan}~\cite{ben2017mutan} and \gls{gmlb}~\cite{minh2019vqa},
we extract image features using \glspl{cnn}.

\subsubsection{Attention Mechanism} Recent breakthroughs have been attributed to attention mechanisms in many \gls{nlp} applications, such as for  neural machine translation~\cite{bahdanau2014neural}, and image classification~\cite{xiao2015application}. Propelled by their remarkable success, numerous \gls{vqa} models employ such schemes to ameliorate predictions as well.

\subsubsection{Fusion Scheme} Current approaches, that fuse image and question features, involve bilinear pooling. The key strength of the bilinear pooling is its ability to enrich the multimodal representations between visual and textual features compared to linear models. Based on the concept of bilinear pooling, Fukui \textit{et al.} developed \gls{mcb}~\cite{fukui2016multimodal} by applying a linear transformation for every pair of image and question features. Kim \textit{et al.} argued that bilinear models like \gls{mcb}, though they provide rich representation, tend to be high-dimensional and limit the applicability to computationally complex tasks~\cite{kim2016hadamard}. Kim \textit{et al.} then proposed the \gls{mlb} algorithm that aimed at reducing the rank of bilinear pooling by utilizing the Hadamard product. A related approach, \gls{mutan}~\cite{ben2017mutan}, from Ben-younes \textit{et al.} efficiently parametrized bilinear interactions between image and question features by using multimodal tensor-based Tucker Decomposition. In another work, instead of using low-rank bilinear pooling by applying an inner product operation, Minh \textit{et al.} proposed a full-rank bilinear transformation, \gls{gmlb}~\cite{minh2019vqa}, that is capable of learning a great range of answers.

Despite achieving impressive results, all aforementioned fusion models rely on an `unweighted' \MV{`balanced'?} multimodal fusion (\textit{i.e.} consider that image and question features are equally important). These fusion schemes may make \gls{vqa} systems pay less attention to either image or question that are supposed to be more critical towards answering the question. In the present work, we propose a \gls{qcmlb} approach that puts more emphasis on the question features.

\section{Proposed VQA Model}
\label{sec:proposed_approach}

\begin{figure*}[!t]
\centering
\includegraphics[width=0.9\textwidth]{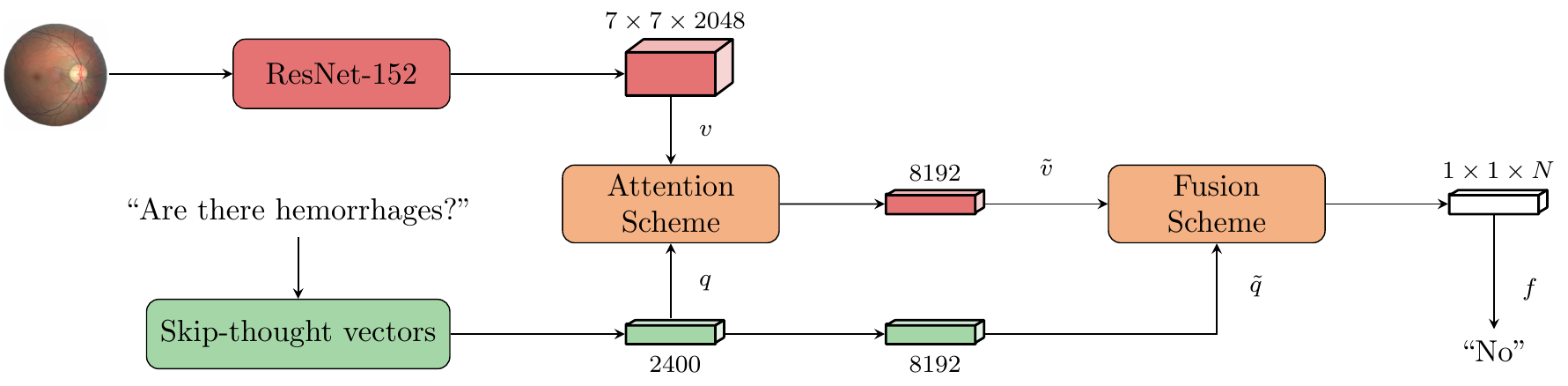}
\caption{The proposed \gls{qcmlb} model together with the attention mechanism. A ResNet-152 and Skip-thought vectors were used to extract image, $v$, and question features, $q$, respectively. These are combined using an attention mechanism in order to produce global image features, $\tilde{v}$, that are then fused with global question features, $\tilde{q}$, to output $N$ answer likelihoods. Here, $J=2400$, $K=2048$ and $G=4$ denote the dimension of image features, dimension of question features, and number of glimpses, respectively.}
\label{fig:vqa_baseline}
\end{figure*}

As with existing \gls{vqa} methods, our goal is to predict the most likely answer, $\hat{a}$, to a question, $q$, about an image, $v$. The problem can be formulated as finding,
\begin{equation} 
    \hat{a} = \argmaxA_{a\in\mathcal{A}} P(a \,|\, q,v, \Theta),
    \label{eqn:vqa}
\end{equation}
where $\mathcal{A}$ is the set of possible answers and $\Theta$ contains all model parameters. 

\figref{fig:vqa_baseline} illustrates the proposed \gls{qcmlb} method. It extracts the image, $v$, and question features, $q$, using pre-trained networks, and applies a fusion scheme to jointly embed the image and question features in the same space. The features are combined using a multi-glimpse attention mechanism~\cite{fukui2016multimodal} and a novel fusion scheme for global image features and global question features. In particular, we propose a simple yet powerful {\it question-centric} scheme that emphasizes the questions. The resulting latent feature vector of dimension $N=|\mathcal{A}|$ is passed through a softmax activation and used to predict the most likely answer. Next, we describe in detail the complete fusion scheme.

\subsection{\gls{qcmlb} Formulation}
\label{subsec:model_formulation}

To efficiently encode the relationships between questions and images, we make use of a multi-glimpse attention mechanism proposed by 
Xu \textit{et al.}~\cite{xu2015show}. The purpose of the multi-glimpse attention mechanism is to allow the \gls{vqa} model to make the input question select appropriate regions of the input image to answer the question. To do this, we first project a spatial grid of locations with respect to the image features into the semantic question space to calculate a weight matrix that corresponds to an attention map correlating the location grid and the input question. Then, instead of learning a single attention map, we utilize a multi-glimpse scheme such that each glimpse represents an attention map. Using the multi-glimpse attention mechanism, global image features are computed as,
\begin{equation} \label{eqn:globalimagefeatures}
  \tilde{v} = [\omega_1^T, \ldots, \omega_G^T]^T \in \mathbb{R}^{KG},
\end{equation}
where the \textit{glimpses} are,
\begin{equation} \label{eqn:glimpse}
  \omega_g
      = \left[q^T\mathbf{W}_{1g}^e v_0 + B^e_{1g}, \ldots, q^T\mathbf{W}_{Kg}^e v_0 + B^e_{Kg} \right]^T,
\end{equation}
whereby
$q \in \mathbb{R}^{J}$ are the question features and $v_0=\text{Vec}(v) \in \mathbb{R}^{K}$ are the vectorized image features, with $J=2400$, $K=2048$, $G$ denoting the number of glimpses, 
and
$\mathbf{W}^e_{kg}\in\mathbb{R}^{J\times K}$ and $B_{kg}^e\in\mathbb{R}$ are the weight matrices and biases, respectively, in the given \textit{attention} scheme (denoted by superscript $e$).

Before concatenating or tiling question features, the question features before must be transformed. Denoted \textit{pre-tiled} question features (see Figure~3 in \cite{fukui2016multimodal}) are computed as,
\begin{equation} 
\label{eqn:pretiled}
    \hat{q} = {\relu}({\mathbf{W}^{\hat{q}} q + b^{\hat{q}}}),
\end{equation}
where $\hat{q} \in \mathbb{R}^{K}$, and $\mathbf{W}^{\hat{q}} \in \mathbb{R}^{K \times J}$ and $b^{\hat{q}} \in \mathbb{R}^{K}$ denote the pre-tiled weight matrix and bias terms, respectively. The global or tiled question features are then defined as,
\begin{equation} \label{eqn:tiledquestion}
    \tilde{q} = [\hat{q}_1, \hat{q}_2, \cdots \hat{q}_G]^T,
\end{equation}
where $\tilde{q} \in \mathbb{R}^{KG}$ and $\hat{q}_1 = \hat{q}_2 = \cdots = \hat{q}_G = \hat{q}$. 

Given these, the output features are encoded using bilinear pooling,
\begin{equation} \label{eqn:bilinear}
  f_n =
      \sum_{j=1}^{KG} \sum_{k=1}^{KG} \tilde{q}_j w^f_{n j k} \tilde{v}_{k} + b^f_{n}
      = \tilde{q}^T \mathbf{W}_n^f \tilde{v} + b^f_{n}, 
\end{equation}
where $f_n$ is the answer feature corresponding to $a_n \in \mathcal{A}$, and
$\mathbf{W}_n^f\in\mathbb{R}^{KG \times KG}$ and $b^f\in\mathbb{R}^N$ denote the weight and bias terms in the \textit{fusion} scheme, respectively. 
Recall that there are $N=|\mathcal{A}|$ possible answers, hence the output feature vector is
\begin{equation} \label{eqn:answer_before_decomposed}
    f = [f_1, \ldots, f_N]^T \in \mathbb{R}^{1\times1\times N}.
\end{equation}

\begin{table}[!t]
    \centering
    \begin{adjustbox}{max width=0.45\textwidth,center}
    \begin{tabular}{l c l l}
        \toprule
        \textbf{Notation}   & & \textbf{Description}        \\ 
        \cmidrule{1-1} \cmidrule{3-3}
        $a$                 & & correct answer              \\
        $\hat{a}$           & & predicted answer            \\
        $q$                 & & question features           \\
        $\tilde{q}$         & & global question features    \\
        $\hat{q}$           & & pre-tiled question features \\
        $v$                 & & image features              \\
        $\tilde{v}$         & & global image features       \\
        $f$                 & & output features             \\
        $e$                 & & attention scheme            \\
        $\mathcal{A}$       & & set of possible answers     \\
        $\Theta$            & & model parameters            \\
        $K$                 & & dimension of image features \\
        $J$                 & & dimension of question features \\
        $G$                 & & number of glimpses          \\
        $N$                 & & number of possible answers  \\
        $\omega_g$          & & glimpse $g$                 \\
        $\mathbf{W}$        & & weight matrix               \\
        $\mathbf{B}$        & & bias matrix                 \\
        $b$                 & & bias vector                 \\
        $\times_t$          & & \textit{t-mode} product between a tensor and a matrix    \\
        $\mathbfcal{T}$     & & full tensor                 \\
        $\mathbfcal{T}_c$   & & identity core tensor        \\
        $\mathbf{W}_{\tilde{q}}$, $\mathbf{W}_{\tilde{v}}$, $\mathbf{W}_o$   & & unitary matrices        \\
        \bottomrule
    \end{tabular}
    \end{adjustbox}
    \caption{List of notations used in this paper.}
    \label{tab:notation}
\end{table}

Note that if we assume that the number of possible answers is $N=1000$, then there are at least $KG \times KG \times N \approx 78$ billion parameters to learn. Two issues arise when learning these parameters: (i) fitting these in memory is infeasible for off-the-shelf \glspl{gpu}, and (ii) the learned \gls{vqa} model would be susceptible to heavily overfit. Due to this, Pirsiavash \textit{et al.}~\cite{pirsiavash2009bilinear} proposed a low-rank bilinear method (or low-rank Tucker Decomposition) to reduce the rank of the weight matrix ($\mathbf{W}_n^f$ in our case) and to reduce the number of parameters.

\begin{figure}[!t]
    \begin{center}
        \includegraphics[width=0.4\textwidth]{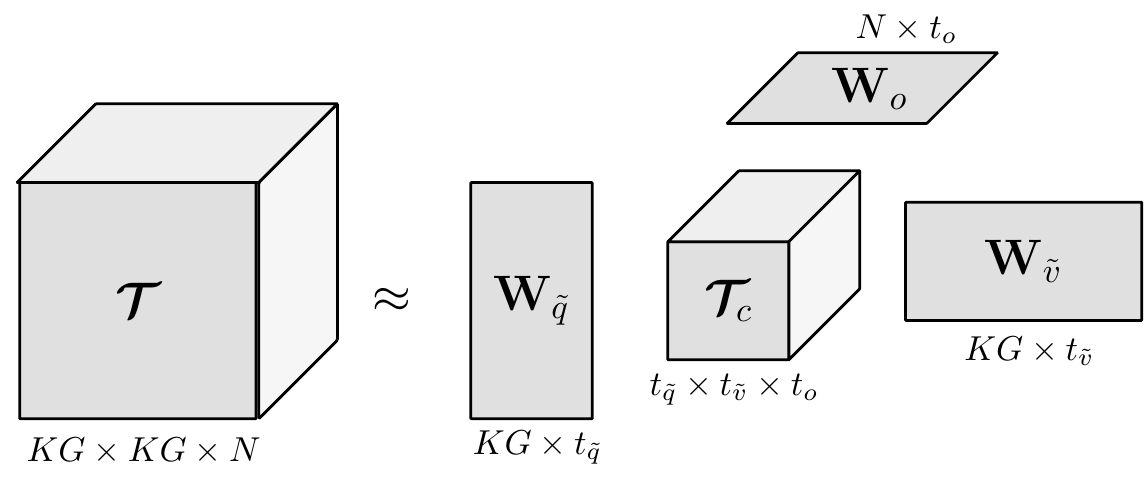}
    \end{center}
    \caption[]{An illustration of Tucker Decomposition.}
    \label{fig:vqa_tucker_decompose}
\end{figure}

Using the Tucker Decomposition~\cite{ben2017mutan}, \eqref{eqn:answer_before_decomposed} can be formulated as,
\begin{equation} \label{eqn:answer}
    f = (\mathbfcal{T}\times_1 \tilde{q}^T) \times_2 \tilde{v}^T + b^f,
\end{equation}
where the operator $\times_t$ denotes the \textit{t-mode} product between a tensor and a matrix. The full tensor, $\mathbfcal{T} \in \mathbb{R}^{KG \times KG \times N}$, can then be further decomposed as,
\begin{equation} \label{eqn:tucker}
    \mathbfcal{T} = \big((\mathbfcal{T}_c \times_1 \mathbf{W}_{\tilde{q}}) \times_2 \mathbf{W}_{\tilde{v}}\big) \times_3 \mathbf{W}_o,
\end{equation}
where
$\mathbfcal{T}_c \in \mathbb{R}^{t_{\tilde{q}} \times t_{\tilde{v}} \times t_o}$
is an identity core tensor with dimensions
$t_{\tilde{q}},t_{\tilde{v}},t_o\in\mathbb{Z}^+$,
and
$\mathbf{W}_{\tilde{q}} \in \mathbb{R}^{KG \times t_{\tilde{q}}}$, $\mathbf{W}_{\tilde{v}} \in \mathbb{R}^{KG \times t_{\tilde{v}}}$, and $\mathbf{W}_o \in \mathbb{R}^{N \times t_o}$ are unitary matrices. As in~\cite{ben2017mutan}, we combine Equations~\ref{eqn:answer} and~\ref{eqn:tucker} such that
\begin{equation} \label{eqn:answerexpand}
    f = \Big(\big(\mathbfcal{T}_c \times_1 (\tilde{q}^T\mathbf{W}_{\tilde{q}})\big)
    \times_2 (\tilde{v}^T\mathbf{W}_{\tilde{v}})\Big) \times_3 \mathbf{W}_o.
\end{equation}

We propose a simple, yet efficient extension, inspired by \gls{mlb}~\cite{kim2016hadamard} and the observation that question-only models perform substantially better than image-only models~\cite{kafle2017visual}. Specifically, we introduce a \textit{question-centric} model that improves the question features by emphasizing them and their part of the model.

Following \eqref{eqn:answerexpand}, the proposed model is expressed by,
\begin{equation} \label{eqn:answerproposed}
    f_{\text{qcmlb}} = \Big(\big(\mathbfcal{T}_c \times_1 (\tilde{q}^T\mathbf{W}_{\tilde{q}})^2\big) \times_2 (\tilde{v}^T\mathbf{W}_{\tilde{v}})\Big) \times_3 \mathbf{W}_o.
\end{equation}
\noindent
Further, the proposed model emphasizes the \textit{pre-tiled} question features such that,
\begin{equation} \label{eqn:pretiledproposed}
    \hat{q}_{\text{qcmlb}} = {\relu}({\mathbf{W}^{\hat{q}} q + b^{\hat{q}}})^2,
\end{equation}
where $\hat{q}_{\text{qcmlb}} \in \mathbb{R}^{K}$, and $\mathbf{W}^{\hat{q}} \in \mathbb{R}^{K \times J}$ and $b^{\hat{q}} \in \mathbb{R}^{K}$ denote the pre-tiled weight matrix and bias terms, respectively. 

Note that, the squares in \eqref{eqn:answerproposed} and \eqref{eqn:pretiledproposed} are computed element-wise and are convenient to explore effective polynomial transformations to emphasize the questions. In the following section, we describe how we do this.

\begin{figure*}[!th]
    \begin{center}
        \includegraphics[width=0.9\textwidth]{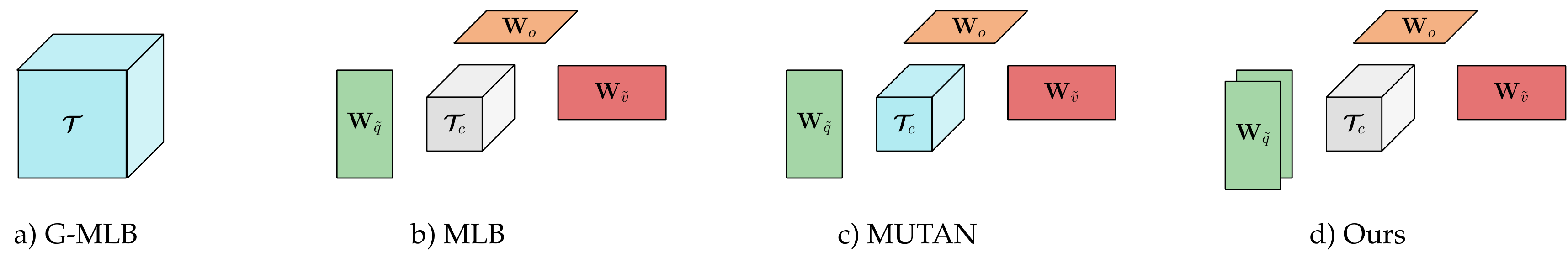}
    \end{center}
    \caption[]{A comparison of \gls{vqa} fusion models. a) \gls{gmlb}~\cite{minh2019vqa}: full tensor $\mathbfcal{T}$ is trainable and not decomposed as in other approaches. b) \gls{mlb}~\cite{fukui2016multimodal}: $\mathbf{W}_{\tilde{q}}$, $\mathbf{W}_{\tilde{v}}$ and $\mathbf{W}_{\tilde{o}}$ are trainable, while $\mathbfcal{T}_c$ is fixed. c) \gls{mutan}~\cite{ben2017mutan}: all four elements are trainable. d) Ours: similar to MLB with proposed element-wise square of $\mathbf{W}_{\tilde{q}}$. Gray color denotes fixed tensor.}
    \label{fig:vqa_compare_fusion}
\end{figure*}

\subsection{Hyperparameter Grid-search}
\label{subsec:grid}

To find a beneficial way to transform the image and question features, we employed a grid search over parameters $m$ and $n$,
\begin{equation} \label{eqn:answergeneral}
    f_{\text{grid}} = \Big(\big(\mathbfcal{T}_c \times_1 (\tilde{q}^T\mathbf{W}_{\tilde{q}})^m\big) \times_2 (\tilde{v}^T\mathbf{W}_{\tilde{v}})^n\Big) \times_3 \mathbf{W}_o,
\end{equation}
where $f_{\text{grid}}$ is the general form of $f$ of \eqref{eqn:answerexpand}.

We employed a grid search over element-wise polynomial functions up to degree three and evaluated each model on an independent validation set (see \tableref{tab:result_type_new}) so to avoid any training bias.

Ultimately, the element-wise square of the global and pre-tiled question features (see \eqref{eqn:answerproposed} and \eqref{eqn:pretiledproposed}) selected was the transformation that performed the best on the independent validation set of all polynomial element-wise transformations evaluated in the grid-search procedure (see \tableref{tab:result_grid}). Experimentally, we found that the square transformations on the question features revealed the ``question-centric'' term in the proposed model's name.

\section{Question-Answer pairs and implementation details }
\label{subsec:experiment}

In this section, we detail our proposed method for \gls{qa} pair generation to train our model. To do this, we specify the different datasets we used in this work as the \gls{qa} will depend on these, as well as the implementation details of our approach.

\subsection{Datasets} \label{subsec:dataset}

The \textit{Indian Diabetic Retinopathy Image Dataset (IDRiD)}\footnote{\url{https://idrid.grand-challenge.org/}}~\cite{porwal2018indian} contains color fundus images of the retina with a resolution of $\text{4,288} \times \text{2,848}$, whereby each of the 516 images contains disease severity levels as well as regions of four retinal biomarkers: \textit{microaneurysms}, \textit{hemorrhages}, \textit{hard exudates}, and \textit{soft exudates} which aims to detect the disease severity level of \gls{dr} (five levels) and \gls{dme} (three levels). As such, this datatset allows for classification and segmentation tasks.

The \textit{BreAst Cancer Histology Dataset (BACH)}\footnote{\url{https://iciar2018-challenge.grand-challenge.org/}}~\cite{aresta2018bach} comprises over 400 labeled microscopy images with ten pixel-wise labeled and 20 non-labeled whole-slide images with an ultra-high resolution of $\text{42,113} \times \text{62,625}$ pixels. The microscopy images were annotated by two expert clinicians into four classes: \textit{normal}, \textit{benign}, \textit{in situ carcinoma}, and \textit{invasive carcinoma} according to the preponderant cancer. This data also allows for both classification and segmentation tasks.

The \textit{Tool Detection and Operative Skill Assessment Dataset (Tools)}\footnote{\url{http://ai.stanford.edu/~syyeung/tooldetection.html}}~\cite{jin2018tool} consists of spatial bounding boxes of $\text{2,532}$ frames across ten real-world laparoscopic surgical videos. With an average of 1.2 labels per frame, this dataset comprises $\text{3,141}$ annotations on seven surgical tools. The tools are: \textit{grasper}, \textit{bipolar}, \textit{hook}, \textit{scissors}, \textit{clip applier}, \textit{irrigator}, and \textit{specimen bag}.

The \textit{Visual Question Answering in the Medical Domain} (\textit{VQA-Med})\footnote{\url{https://www.imageclef.org/2019/medical/vqa}}~\cite{ImageCLEFVQA-Med2019} contains a total of $\text{4,200}$ images with $\text{15,292}$ corresponding \gls{qa} pairs, and is partitioned in three sets: training, validation, and test sets. The validation and test sets both contain $\text{500}$ images, and the training set thus contains $\text{3,200}$ images. The training set consists of $\text{12,792}$ \gls{qa} pairs, the validation set $\text{2,000}$, and the test set consists of $\text{500}$ \gls{qa} pairs. The questions are categorized into four types, with several varieties: Modality (36), Plane (16), Organ System (10), and Abnormality. The task of the VQA-Med challenge is to predict the most likely answer given the medical images. Apart from the strict accuracy, the \gls{bleu} score was also used to evaluate in the 2019 \textit{VQA-Med} challenge.

Finally, the two \textit{VQA Datasets}\footnote{\url{https://visualqa.org/}} contain open-ended questions about images from the MS-COCO natural image dataset\footnote{\url{http://cocodataset.org/}}~\cite{lin2014microsoft}, where each image has at least three questions. Each question is paired with ten answers from unique specialists. Roughly $40~\%$ of the questions have a ``yes'' or ``no'' answer. The two available versions of this dataset are: (1) \textit{VQA-V1} with $\text{204,721}$ MS-COCO images with $\text{614,163}$ questions and $\text{50,000}$ abstract images with $\text{150,000}$ questions; and (2) \textit{VQA-V2} with $\text{265,016}$ MS-COCO images coupled with $\text{1,105,904}$ questions.

\subsection[QA-Pair Ground Truth Generation]{Question and Answer (QA) Pair Generation}
\label{subsec:qagen}

\begin{table*}[!th]
    \centering
    \begin{tabular}{l c l C{1.2cm} C{1.2cm} C{1.2cm}}
        \toprule
        \textbf{Question types}                         && \textbf{answer}  & \textbf{IDRiD} & \textbf{BACH} & \textbf{Tools} \\ 
        \cmidrule{1-1} \cmidrule{3-6}
        is there any $x$                                && yes/no           & \checkmark & \checkmark   & \checkmark    \\ 
        is $x$ larger/smaller than $y$                  && yes/no/n.a.      & \checkmark & \checkmark   &               \\
        is $x$ in $z$                                   && yes/no           & \checkmark & \checkmark   & \checkmark    \\ 
        how many classes/tumors are there               && number           &            & \checkmark   &               \\ 
        what is the major/minor class/tumor             && name             &            & \checkmark   &               \\ 
        how many pixels/percent of $x$                  && number           &            & \checkmark   &               \\ 
        how many tools are there                        && number           &            &              & \checkmark    \\ 
        which tool has pointed tip position on $t$      && position         &            &              & \checkmark    \\
        \cmidrule{1-1} \cmidrule{3-6}
        \multicolumn{2}{l}{\textbf{Number of QA pairs}} &                   & 220~k      & 360~k        & 1~M           \\
        \bottomrule
    \end{tabular}
    \caption{\gls{qa} pairs used for the datasets IDRiD, BACH and Tools. $x$ and $y$ denote disease classes, such as \textit{benign} or \textit{in situ carcinoma} in BACH, $z$ encodes a specific image location, and $t$ denotes an image region.}
    \label{tab:qa}
\end{table*}

In this section, we describe how we automated the \gls{qa} pair generation with minimal supervision using typical datasets found in the medical image computing literature. \tableref{tab:qa} lists the types of questions used in this work for the IDRiD, BACH, and Tools datasets after thorough discussions with physicians about their concerns when examining medical images. Eleven question types and four answer types (``yes'' or ``no'', numbers, categories, and locations) were used. These questions implicitly query challenging sub-tasks, such as classification and identification (``is there any $x$''), counting (``how many classes''), semantic segmentation (``is $x$ larger than $y$''), localization (``is there any $x$ in $z$''), object detection (``how many classes/tumors are there'', ``how many tools are there''), image understanding (``which tool has pointed tip position to the ``$t$''), and image quantification (``how many pixels/percent of $x$''). Here, $x$ and $y$ denote a disease or a class, locations are defined as either general directions ($t$, \textit{e.g.} top, bottom, left, right), or as rectangles ($z$, \textit{e.g.} pixels (0, 0) to (32, 32)). This allows questions to be asked about locations in the image. \gls{qa} pairs were generated for each dataset using the question templates found in \tableref{tab:qa}. Of these, 80~\% of the questions were used for training and 20~\% for testing.

We now explain how each type of question was generated. First, ``is there any $x$'' is a classification task where the answers are either ``yes'' or ``no.'' An example of this type of question is ``are there microaneurysms in the fundus?'' Generating this type of QA groundtruth is straight-forward: we counted the number of non-zero pixels microaneurysms in a groundtruth image. If this number was greater than 0, we marked the answer as ``yes'', if not, we marked it as ``no''.

Second, ``are the hard exudates larger than the microaneurysms?'' is an example of question type ``is $x$ larger than $y$''. For this type of question, we also counted the number of non-zero pixels in hard exudates and microaneurysms in the corresponding groundtruth images. If the hard exudates had more non-zero pixels than the microaneurysms, we labeled the answer as ``yes''; otherwise, the answer was ``no''. Different from the first kind of question, the second type comprises another possible answer that is ``n.a.'' or ``undefined''. ``Undefined'' was chosen when there were neither hard exudates nor microaneurysms in the image.

Third, ``is there any $x$ in $z$?'' is a localization task that demands the \gls{vqa} model to be able to localize an object and decoding a location. An example can be ``are there any microaneurysms in pixel locations (0, 0) to (32, 32)?'' Here, (0, 0) and (32, 32) denote the $(x, y)$ coordinates of the corners of the region of interest. The answer was ``yes'' if there was ``x'' in the aforementioned location, and ``no'', otherwise. 

Fourth, ``how many classes/tumors are there?'' or ``how many tools are there?'' is an object counting task. For this type of question, the answers were generated by computing the number of classes/tumors/tools in the image.

Fifth, ``which tool has pointed tip position to the $t$?'' is an image understanding problem. In such a task, the \gls{vqa} algorithm is required to not only classify objects successfully by answering ``which objects are there in the image?'', but also locate these objects correctly. To generate the groundtruth answer, we first identified the objects, that appeared in the image, and then found their corresponding locations. There was always at least one tool in an image in the Tools dataset, hence there were no ``undefined'' answers.

Last, ``what is the major/minor class/tumor?'' and ``how many pixels/percent of $x$?'' defines a quantification task that requires the \gls{vqa} model to be able to classify, segment, and quantify the area of every single tumor/tool/object in an image. To generate the answer, we did as follows: (i) detecting non-background objects, (ii) segmenting these found objects, (iii) counting the number of non-zero pixels in the segmentation mask belonging to each object, and (iv) returning a relevant groundtruth answer.

\subsection{Implementation details and training}
\label{subsec:implement}

The proposed method was implemented using the PyTorch library~\cite{paszke2017automatic}
and the experiments were run on GTX 1080 Ti 11~GB and NVIDIA Tesla V100 16~GB GPUs. The implementation is available at \url{https://github.com/vuhoangminh/vqa_medical}.

The proposed \gls{qcmlb} model, illustrated in \figref{fig:vqa_baseline}, contains three different components: an image model, a question model, and the proposed fusion with an attention mechanism model. Each model was trained separately as described below.

\subsubsection{Image Model} 
\label{subsubsec:image} 
We used the ResNet-152 network to extract visual features. We used the pre-processing pipeline described in~\cite{minh2019vqa} for the VQA-Med dataset, but the other datasets (see \secref{subsec:dataset}) were only pre-processed by resizing the input images to $448 \times 448$ and $Z$-normalizing them.

The networks for the medical images were trained from their random initialization with the Adam optimizer~\cite{kingma2014adam}, an initial learning rate of $0.0001$ and momentum parameters of $\beta_1=0.9$ and $\beta_2=0.999$. The mini-batch size was four, and we used early stopping to handle overfitting: the training was aborted if the accuracy on the validation set did not improve over 10 iterations. Note that, the ResNet-152 models were trained from their initial configuration on classification tasks for the BACH, IDRiD, and Tools datasets (see \secref{subsec:dataset}), while a pre-trained ResNet-152 (on the ImageNet dataset\footnote{\url{https://github.com/facebook/fb.resnet.torch}}) was used for the VQA-Med, VQA-V1, and VQA-V2 datasets. These aforementioned classification tasks were either generated (Tools) or taken from the classification challenges introduced in BACH and IDRiD. Note that, an alternative to extract image features for the BACH, IDRiD, and Tools datasets would also be a pre-trained on the ImageNet ResNet-152 as used in the VQA-Med, VQA-V1, and VQA-V2 datasets.

\subsubsection{Question Model} 
\label{subsubsec:ques} 
Questions were pre-processed by removing the punctuation marks and converting them to lower-case, as in~\cite{ben2017mutan,fukui2016multimodal}. Skip-thought vectors were pre-trained on the BookCorpus dataset~\cite{zhu2015aligning} with $\text{1,316,420}$ unique words in 16 different genres~\cite{kiros2015skip} to obtain pre-trained Skip-thought vectors, $\mathbfcal{Q}_{st}$. The lack of medical terms in the BookCorpus prevents the Skip-thought vectors to differentiate between unknown or unseen words, for example, \textit{microaneurysms} or \textit{hemorrhages}, that were used in the Diabetic Retinopathy Image Dataset. 

To overcome this shortcoming, we use the transfer learning approach introduced in~\cite{kiros2015skip} to encode words that may not have been seen before. First, we used a Word2Vec model~\cite{mikolov2013distributed}, $\mathbfcal{Q}_{w2v}$, which was trained on the Google News dataset and which contains word vectors for 3 million words and phrases~\cite{mikolov2013efficient}. Second, a linear regression model, with weights $\mathbf{W}$, was fit to map the Word2Vec embedding space, $q_{w2v}$, to the Skip-thought embedding space, $q_{st}$, by minimizing a least-squares criterion. Finally, Skip-thought question features were generated by computing,
\begin{equation}
    \widehat{q}_{st} = \mathbf{W}q_{w2v}.
\end{equation}
\noindent
This mapping then allowed word features to cover both natural and medical domain questions.

\subsubsection{Fusion Model and the Attention Mechanism} 
To implement \eqref{eqn:answer} and \eqref{eqn:tucker}, we followed the description in \cite{ben2017mutan}. We employed the Adam optimizer with early stopping, a learning rate of 0.0001 with corresponding exponential decay rates of $\beta_1=0.9$ and $\beta_2=0.999$, and a mini-batch size of 128 when using the attention scheme and a mini-batch size of 512 without the attention scheme. 

During training, the loss function was the categorical cross-entropy, while the evaluation metric was the strict accuracy. In addition to the strict accuracy, precision and recall were used in the performance evaluation process. We used a dropout rate of 0.5 for all dense layers. Using an Nvidia GTX 1080 Ti GPU, the training time for the networks without attention was about 6 hours and with the attention scheme, it was more than two days.


\section{Experiments and Results} \label{subsec:result}

In this section, we compare our \gls{qcmlb} model to other recent approaches under 
two schemes: with and without the attention mechanism. In particular, we compare our
approach to \gls{mlb}~\cite{kim2016hadamard}, \gls{mutan}~\cite{ben2017mutan} and \gls{gmlb}~\cite{minh2019vqa}. The \gls{mlb} and \gls{mutan} 
models were trained using the code available from the \gls{mutan} Github page\footnote{\url{https://github.com/Cadene/vqa.pytorch}}. The evaluation metrics were strict macro-accuracy, defined as the percentage of correct answers, macro-precision and macro-recall scores for all datasets mentioned in \secref{subsec:dataset}. For the sake of readability, we will denote macro-accuracy, macro-precision, and macro-recall by just accuracy, precision, and recall from here on.

\subsection{Quantitative Results} \label{subsec:quantitative_result}

\tableref{tab:result_grid} shows the mean accuracy for all types of questions (computed from the last 21 epochs) for a subset of the polynomial degrees tested (up to degree three), and for three of the datasets (the VQA-V1, VQA-V2, and IDRiD) using the ``Ours + tanh + Att'' model. From \tableref{tab:result_grid}, we see that the element-wise square transformation performs the best on the three aforementioned datasets.

\begin{table}[!th]
    \centering
    \begin{adjustbox}{max width=0.4\textwidth}
    \begin{tabular}{cc c ccc}
        \toprule
        m & n &&\textbf{IDRiD}&\textbf{VQA-V1}& \textbf{VQA-V2} \\
        \cmidrule{1-2} \cmidrule{4-6}
        3 & 1 & & 88.31 (0.32)   & 53.78 (0.10)   & 50.72 (0.10)\\
        2 & 1 & & \textbf{89.38 (0.40)}   
                                 & \textbf{54.03 (0.09)} 
                                                  & \textbf{51.31 (0.14)}\\
        1 & 1 & & 89.05 (0.41)   & 53.63 (0.07)   & 51.06 (0.08)\\
        1 & 2 & & 87.65 (0.35)   & 53.18 (0.08)   & 50.65 (0.09)\\
        1 & 3 & & 88.41 (0.29)   & 53.02 (0.11)   & 50.81 (0.12)\\
        \bottomrule
    \end{tabular}
    \end{adjustbox}
    \caption{Mean macro-accuracy (and standard errors) of all types of question computed from the last 21 epochs for a subset of the polynomial degrees tested (up to degree three), and for three of the datasets (the IDRiD, VQA-V1 and VQA-V2 datasets) using the ``Ours + tanh + Att'' model. 
    Here, $m$ and $n$ are the polynomial degrees for the question and image features, respectively (see \eqref{eqn:answergeneral}).}
    \label{tab:result_grid}
\end{table}

\begin{table}[!th]
    \def\width{2 cm}
    \centering
    \begin{adjustbox}{max width=0.48\textwidth}
    \begin{tabular}{rccccccccc}
        &
        \rotso{$\text{MUTAN}$ + tanh \cite{ben2017mutan}} &
        \rotso{$\text{MLB}$ + tanh \cite{fukui2016multimodal}} &
        \rotso{$\text{MUTAN}$ + tanh + Att \cite{ben2017mutan}} & 
        \rotso{$\text{MLB}$ + tanh + Att \cite{fukui2016multimodal}} &
        \rotso{$\text{G-MLB} + \relu$ + Att \cite{minh2019vqa}} &
        \rotso{$\text{Ours}$ + tanh} &
        \rotso{$\text{Ours}$ + tanh + Att} &
        \rotso{$\text{Ours} + \relu$} &
        \rotso{$\text{Ours} + \relu$ + Att} \\
        \cmidrule{1-10}
        $\text{MUTAN}$ + tanh \cite{ben2017mutan}           & ~~~ & $0$ & $-$ & $-$ & $-$ & $0$ & $-$ & $-$ & $-$  \\
        $\text{MLB}$ + tanh \cite{fukui2016multimodal}      & $0$ & ~~~ & $-$ & $-$ & $-$ & $0$ & $-$ & $-$ & $-$  \\
        $\text{MUTAN}$ + tanh + Att \cite{ben2017mutan}     & $+$ & $+$ & ~~~ & $-$ & $-$ & $+$ & $-$ & $0$ & $-$  \\
        $\text{MLB}$ + tanh + Att \cite{fukui2016multimodal}& $+$ & $+$ & $+$ & ~~~ & $0$ & $+$ & $-$ & $+$ & $-$  \\
        $\text{G-MLB} + \relu$ + Att \cite{minh2019vqa}     & $+$ & $+$ & $+$ & $0$ & ~~~ & $+$ & $0$ & $+$ & $-$  \\
        $\text{Ours}$ + tanh                                & $0$ & $0$ & $-$ & $-$ & $-$ & ~~~ & $-$ & $-$ & $-$  \\
        $\text{Ours}$ + tanh + Att                          & $+$ & $+$ & $+$ & $+$ & $0$ & $+$ & ~~~ & $+$ & $0$  \\
        $\text{Ours} + \relu$                               & $+$ & $+$ & $0$ & $-$ & $-$ & $+$ & $-$ & ~~~ & $-$  \\
        $\text{Ours} + \relu$ + Att                         & $+$ & $+$ & $+$ & $+$ & $+$ & $+$ & $0$ & $+$ & ~~~  \\
        \bottomrule
    \end{tabular}
    \end{adjustbox}
    \caption{The results of the Nemenyi post-hoc test comparing all evaluated methods. A minus ($-$) means ranked significantly lower, a zero ($0$) means non-significant difference, and a plus ($+$) means ranked significantly higher, when comparing a method in the rows to a method in the columns.}
    \label{tab:result_nemenyi}
\end{table}

\begin{table*}[!th]
    \def\width{0.82 cm}
    \def\widthdetail{1.5 cm}
    \centering
    \begin{adjustbox}{max width=0.94\textwidth,center}
    \begin{tabular}{rrrrc C{\width}C{\width}C{\width}C{\width}|C{\width}|C{\width}C{\width}|C{\width}C{\width}}
    & 
    \rot{question type} & 
    \rot{proportion} &
    \rot{\# possible answers} &
    & 
    \rot{$\text{MUTAN}$ + tanh \cite{ben2017mutan}} &
    \rot{$\text{MLB}$ + tanh \cite{fukui2016multimodal}} &
    \rot{$\text{MUTAN}$ + tanh + Att \cite{ben2017mutan}} & 
    \rot{$\text{MLB}$ + tanh + Att \cite{fukui2016multimodal}} &
    \rot{$\text{G-MLB} + \relu$ + Att \cite{minh2019vqa}} &
    \rot{$\text{Ours}$ + tanh} &
    \rot{$\text{Ours}$ + tanh + Att} &
    \rot{$\text{Ours} + \relu$} &
    \rot{$\text{Ours} + \relu$ + Att} \\
    \toprule
    \multirow{7}{*}{\textbf{BACH}}      & yes/no          & 95.62 \% & 3         && 89.97 (0.24) & 89.55 (0.15) & 89.96 (0.13) & 90.35 (0.08) & 90.25 (0.07) & 89.56 (0.14) & 90.47 (0.05) & 90.37 (0.28) & 90.90 (0.03) \\ 
                                        & number          & 3.37  \% & 65,536    && 81.38 (0.08) & 80.07 (0.12) & 81.07 (0.15) & 81.22 (0.14) & 81.45 (0.11) & 80.48 (0.07) & 81.60 (0.07) & 81.08 (0.21) & 81.88 (0.14) \\ 
                                        & name            & 1.01  \% & 5         && 80.25 (0.45) & 80.51 (0.19) & 81.76 (0.26) & 81.31 (0.13) & 81.95 (0.15) & 81.27 (0.19) & 81.54 (0.10) & 80.69 (0.49) & 82.62 (0.14) \\ 
    \cmidrule{2-4} \cmidrule{6-14}
                                        & all             & 100.00 \%& 65,544    && 90.48 (0.25) & 90.03 (0.45) & 90.48 (0.17) & 90.85 (0.10) & 90.77 (0.20) & 90.07 (0.23) & 90.99 (0.10) & 90.86 (0.21) & \textbf{91.42 (0.27)} \\ 
    \cmidrule{1-14}
    \multirow{3}{*}{\textbf{IDRiD}}     & yes/no          & 100.00 \%& 3         && 86.55 (0.25) & 85.95 (0.41) & 88.85 (0.64) & 89.05 (0.41) & 90.30 (0.34) & 88.31 (0.23) & 89.38 (0.40) & 88.13 (0.56) & 89.68 (0.31) \\ 
    \cmidrule{2-4} \cmidrule{6-14}
                                        & all             & 100.00\% & 3         && 86.55 (0.25) & 85.95 (0.41) & 88.85 (0.64) & 89.05 (0.41) & \textbf{90.30 (0.34)} & 88.31 (0.23) & 89.38 (0.40) & 88.13 (0.56) & 89.68 (0.31) \\ 
    \cmidrule{1-14}
    \multirow{5}{*}{\textbf{Tools}}     & yes/no          & 98.91 \% & 2         && 96.26 (0.01) & 96.30 (0.01) & 96.34 (0.02) & 96.40 (0.02) & 96.67 (0.02) & 96.32 (0.01) & 96.46 (0.01) & 96.42 (0.01) & 96.68 (0.02) \\ 
                                        & position        & 0.87  \% & 8         && 70.16 (0.30) 
                                                                                                 & 66.00 (0.27) & 67.36 (0.82) & 67.14 (1.05) & 64.86 (0.99) & 63.41 (0.38) & 64.52 (1.13) & 65.88 (0.77) & 67.35 (0.59) \\ 
    \cmidrule{2-4} \cmidrule{6-14}                                        
                                        & all             & 100.00\% & 10  && 95.97 (0.05) & 95.96 (0.03) & 96.01 (0.11) & 96.09 (0.06) & 96.32 (0.05) & 95.94 (0.03) & 96.11 (0.06) & 96.07 (0.04) & \textbf{96.35 (0.04)} \\
    \cmidrule{1-14}
    \multirow{9}{*}{\textbf{VQA-Med}}   & modality        & 25.00 \% & 36        && 82.05 (0.15) & 82.54 (0.11) & 81.81 (0.15) & 79.71 (0.14) & 83.44 (0.12) & 83.02 (0.14) & 83.46 (0.12) & 82.39 (0.14) & 84.28 (0.12) \\ 
                                        & plane           & 25.00 \% & 16        && 72.15 (0.11) & 72.60 (0.12) & 73.57 (0.23) & 75.68 (0.18) & 76.56 (0.17) & 73.50 (0.12) & 73.60 (0.16) & 73.12 (0.15) & 75.65 (0.16) \\ 
                                        & organ system    & 25.00 \% & 10        && 73.20 (0.15) & 73.29 (0.14) & 71.62 (0.21) & 72.13 (0.14) & 74.01 (0.13) & 72.83 (0.10) & 70.97 (0.17) & 73.81 (0.15) & 74.56 (0.18) \\ 
                                        & abnormality     & 25.00 \% & 1,638     && 6.48 (0.14)  & 5.16 (0.04)  & 6.39 (0.08)  & 7.45 (0.06)  & 6.47 (0.07)  & 4.89 (0.05)  & 9.54 (0.09)  & 5.81 (0.08)  & 6.83 (0.10)  \\
    \cmidrule{2-4} \cmidrule{6-14}
                                        & all             & 100.00 \%& 1,700     && 58.47 (0.41) & 58.40 (0.18) & 58.35 (0.18) & 58.74 (0.12) & 60.12 (0.17) & 58.56 (0.25) & 59.39 (0.14) & 58.78 (0.29) & \textbf{60.33 (0.19)} \\ 
    \cmidrule{1-14}
    \multirow{7}{*}{\textbf{VQA-V1}}    & yes/no          & 38.37 \% & 2         && 71.26 (0.15) & 71.47 (0.12) & 72.12 (0.10) & 72.38 (0.03) & 71.77 (0.11) & 70.68 (0.16) & 72.95 (0.07) & 71.38 (0.09) & 72.63 (0.07) \\ 
                                        & number          & 12.31 \% & 8,001     && 30.36 (0.25) & 30.21 (0.15) & 31.46 (0.08) & 30.74 (0.15) & 31.11 (0.21) & 29.86 (0.08) & 31.06 (0.15) & 30.46 (0.16) & 30.92 (0.11) \\ 
                                        & other           & 49.32 \% & 97,978    && 40.37 (0.18) & 40.21 (0.07) & 44.25 (0.16) & 44.76 (0.14) & 41.10 (0.15) & 40.37 (0.19) & 45.05 (0.14) & 40.64 (0.20) & 44.84 (0.08) \\
    \cmidrule{2-4} \cmidrule{6-14}
                                        & all             & 100.00 \%& 105,981   && 50.99 (0.17) & 50.97 (0.06) & 53.37 (0.10) & 53.63 (0.07) & 51.64 (0.10) & 50.71 (0.10) & \textbf{54.03 (0.09)} & 51.18 (0.10) & 53.79 (0.09) \\ 
    \cmidrule{1-14}
    \multirow{7}{*}{\textbf{VQA-V2}}    & yes/no          & 38.37 \% & 2         && 66.12 (0.27) & 65.87 (0.35) & 66.35 (0.24) & 68.91 (0.07) & 69.51 (0.09) & 64.82 (0.37) & 69.28 (0.14) & 66.42 (0.15) & 69.33 (0.02) \\ 
                                        & number          & 12.31 \% & 8,001     && 28.17 (0.40) & 27.85 (0.20) & 28.95 (0.30) & 29.27 (0.04) & 30.13 (0.21) & 27.38 (0.67) & 29.50 (0.27) & 28.34 (0.18) & 29.52 (0.10) \\ 
                                        & other           & 49.32 \% & 97,978    && 37.45 (0.34) & 37.06 (0.36) & 40.71 (0.29) & 42.61 (0.13) & 39.80 (0.19) & 37.02 (0.54) & 42.78 (0.10) & 37.81 (0.12) & 42.81 (0.03) \\
    \cmidrule{2-4} \cmidrule{6-14}
                                        & all             & 100.00 \%& 105,981   && 47.31 (0.31) & 46.98 (0.37) & 49.10 (0.27) & 51.06 (0.08) & 50.01 (0.12) & 46.50 (0.51) & 51.31 (0.14) & 47.62 (0.11) & \textbf{51.35 (0.04)} \\ 
    \bottomrule
    \end{tabular}
    \end{adjustbox}
    \caption{Mean macro-accuracy (and standard errors) of all types and per-type of question (see \tableref{tab:qa}) computed from the last 21 epochs for each method on each dataset. ``+Att'' denotes a method with the attention scheme. $\relu$ and ``tanh'' denote the rectified linear unit and hyperbolic tangent activation functions, respectively.}
    \label{tab:result_type_new}
\end{table*}

\begin{table*}[!th]
    \def\width{2 cm}
    \centering
    \begin{adjustbox}{max width=0.94\textwidth}
    \begin{tabular}{l c C{\width} C{\width} C{\width} C{\width} C{\width} C{\width}}
        \toprule
        \textbf{Model$\backslash{}$Dataset}                 && \textbf{BACH}    & \textbf{IDRiD}    & \textbf{Tools}    & \textbf{VQA-Med}  & \textbf{VQA-V1}   & \textbf{VQA-V2}   \\
        \cmidrule{1-8}
        \multicolumn{8}{l}{\textit{Precision}}\\
        \cmidrule{1-1} \cmidrule{3-8}
        $\text{MUTAN}$ + tanh \cite{ben2017mutan}           && 71.03 (0.29)     & 50.83 (0.14)      & \textbf{60.18 (1.26)}      
                                                                                                                        & 11.46 (0.13)      & 25.07 (0.07)      & 23.03 (0.06)      \\
        $\text{MLB}$ + tanh \cite{fukui2016multimodal}      && 70.72 (0.46)     & 51.49 (0.24)      & 60.06 (0.49)      & 14.52 (0.29)      & 25.22 (0.08)      & 24.18 (0.09)      \\
        $\text{MUTAN}$ + tanh + Att \cite{ben2017mutan}     && 70.99 (0.21)     & 59.50 (0.25)      & 55.48 (1.31)      & 10.86 (0.08)      & 26.03 (0.04)      & 23.23 (0.04)      \\
        $\text{MLB}$ + tanh + Att \cite{fukui2016multimodal}&& 70.23 (1.11)     & 62.11 (0.27)      & 54.70 (1.03)      & 11.57 (0.07)      & 26.02 (0.07)      & 24.77 (0.09)      \\
        \cmidrule{1-1} \cmidrule{3-8}
        $\text{G-MLB} + \relu$ + Att \cite{minh2019vqa}     && 71.13 (0.72)     & 58.71 (0.15)      & 54.64 (1.23)      & 14.89 (0.20)      & \textbf{27.72 (0.14)}         
                                                                                                                                                                & \textbf{26.31 (0.12)}      
                                                                                                                                                                                    \\
        \cmidrule{1-1} \cmidrule{3-8}
        $\text{Ours}$ + tanh                                && 70.74 (0.48)     & 56.09 (0.10)      & 58.29 (0.48)      & \textbf{14.97 (0.27)}      
                                                                                                                                            & 25.66 (0.07)      & 23.53 (0.09)      \\
        $\text{Ours}$ + tanh + Att                          && 70.13 (0.65)     & 64.40 (0.19)      & 53.03 (0.91)      & 10.93 (0.12)      & 25.81 (0.10)      & 24.02 (0.09)      \\
        \cmidrule{1-1} \cmidrule{3-8}
        $\text{Ours} + \relu$                               && 71.27 (0.63)     & 57.35 (0.28)      & 58.05 (0.93)      & 11.33 (0.09)      & 24.81 (0.05)      & 23.78 (0.07)      \\
        $\text{Ours} + \relu$ + Att                         && \textbf{71.27 (0.31)}     
                                                                                & \textbf{64.70 (0.10)}      
                                                                                                    & 53.72 (0.7 )      & 11.24 (0.08)      & 25.90 (0.08)      & 23.49 (0.05)      \\
        \cmidrule{1-8}
        \multicolumn{8}{l}{\textit{Recall}}\\
        \cmidrule{1-1} \cmidrule{3-8}
        $\text{MUTAN}$ + tanh \cite{ben2017mutan}           && 4.98 (0.02)      & 60.53 (0.11)      & 30.57 (0.34)      & 8.60 (0.13)       & 5.09 (0.02)       & 4.86 (0.02)      \\
        $\text{MLB}$ + tanh \cite{fukui2016multimodal}      && 4.92 (0.02)      & 55.61 (0.18)      & 30.68 (0.11)      & 5.52 (0.05)       & 4.92 (0.03)       & 4.35 (0.02)      \\
        $\text{MUTAN}$ + tanh + Att \cite{ben2017mutan}     && 5.03 (0.02)      & 69.00 (0.37)      & 35.30 (0.46)      & 9.17 (0.06)       & 5.54 (0.02)       & 4.99 (0.03)      \\
        $\text{MLB}$ + tanh + Att \cite{fukui2016multimodal}&& 6.10 (0.02)      & 65.93 (0.23)      & 36.17 (0.60)      & 8.59 (0.05)       & 5.60 (0.02)       & 5.49 (0.03)      \\
        \cmidrule{1-1} \cmidrule{3-8}
        $\text{G-MLB} + \relu$ + Att \cite{minh2019vqa}     && 5.14 (0.02)      & \textbf{69.46 (0.16)}      
                                                                                                    & 36.44 (0.52)      & 5.75 (0.04)       & 2.82 (0.03)       & 2.87 (0.02)      \\
        \cmidrule{1-1} \cmidrule{3-8}
        $\text{Ours}$ + tanh                                && 4.93 (0.01)      & 51.47 (0.11)      & 32.43 (0.13)      & 5.12 (0.04)       & 4.33 (0.04)       & 3.93 (0.03)      \\
        $\text{Ours}$ + tanh + Att                          && \textbf{6.31 (0.02)}      
                                                                                & 68.48 (0.23)      & \textbf{37.85 (0.59)}      
                                                                                                                        & 9.45 (0.08)       & \textbf{6.15 (0.01)}       
                                                                                                                                                                & 5.67 (0.02)      \\
        \cmidrule{1-1} \cmidrule{3-8}
        $\text{Ours} + \relu$                               && 5.07 (0.04)      & 67.48 (0.22)      & 32.79 (0.14)      & 8.84 (0.08)       & 5.47 (0.01)       & 4.89 (0.01)      \\
        $\text{Ours} + \relu$ + Att                         && 5.54 (0.01)      & 65.70 (0.15)      & 37.39 (0.56)      & \textbf{9.46 (0.08)}       
                                                                                                                                            & 6.11 (0.02)       & \textbf{6.07 (0.01)}      
                                                                                                                                                                                    \\        
        \bottomrule
    \end{tabular}
    \end{adjustbox}
    \caption{Mean macro-precision and macro-recall of all possible answers/classes computed from the last 21 epochs for each method on each dataset. ``+Att'' denotes a method with the attention scheme. $\relu$ and ``tanh'' denote the rectified linear unit and hyperbolic tangent activation functions, respectively.}
    \label{tab:result_prec_recall}
\end{table*}

\tableref{tab:result_type_new} presents the mean accuracy for all types and per-type of question (computed from the last 21 epochs) for each dataset's validation set. Here we show the performance computed from the last 21 epochs as such statistics are typically noisy when computed from an individual. To report noise-free statistics, and also to avoid accidentally cherry-picking the number of epochs, we average the last few computed statistics at convergence to get a better estimate of the expectation of the statistic. Based on initial exploratory experiments for each model on separate datasets, we found that 21 was effective at representing the convergence of the training process.


To compare the performance of the evaluated methods, we performed a Friedman test of equivalence between the methods, which reported significant differences, and followed it up by a Nemenyi post-hoc test (as proposed in~\cite{Demsar2006}) on all question types with the 21 validation accuracies collected from the last 21 epochs of training (see \tableref{tab:result_nemenyi}). From Table IV and Table V, we see that: (1) \gls{qcmlb} with the $\text{tanh}$ activation function performs better than, or as well as state-of-the-art methods with attention schemes on all evaluated datasets, (2) the attention mechanism boosts performance across all datasets---in particular when comparing \gls{qcmlb} to \gls{mlb}, and (3) \gls{qcmlb} with attention mechanism with the $\relu$ activation function performed significantly better than all other methods, except when comparing to the \gls{qcmlb} with attention mechanism with the tanh activation function, against which there was no significant difference. However, the \gls{qcmlb} with attention mechanism with the tanh activation function also performed with no significant difference compared to the G-MLB with attention mechanism and $\relu$ activation, leaving the proposed \gls{qcmlb} with attention mechanism with the $\relu$ activation function as the highest performing method among those tested.

Since the question-centric extension can be isolated, part of the improved results can be directly attributed to this extension. We also see that the attention mechanism, that emphasizes more important image regions based on this question-centric approach, further improves results. The large performance difference between medical and natural image datasets is likely due to the much larger diversity among images and questions in VQA-V1 and VQA-V2 compared to the medical image datasets. Furthermore, most of the questions in the medical datasets are of type ``yes/no'', and those are likely simpler questions to answer than more open-ended questions found in VQA or VQA-Med datasets (see \tableref{tab:result_type_new}). Indeed, the number of possible answers is much larger in the VQA or VQA-Med datasets (see the number of the possible answer in ~\tableref{tab:result_type_new}) resulting in more challenging inferences. The number of possible answers in ~\tableref{tab:result_type_new} is also the reason why there is a large difference in the performances between yes/no and the other types of questions, especially abnormality in the VQA-Med dataset. However, it is apparent that all models achieve high accuracy for the abnormality question types (the lowest is 4.89~\%) when comparing to the accuracy of randomly selecting the answers.

An explanation as to why the proposed \textit{question-centric} model with the tanh activation function works better than \gls{mlb} and \gls{mutan} could be that the element-wise square operations on the global and pre-tiled question features imply a non-negativity constraint on the latent spaces. This can reduce overfitting and thus improve model generalization. The $\relu$ activation functions further improves the performance that makes ``\gls{qcmlb} + $\relu$ + Att'' the best in most tasks. This would be in line with the well-documented positive $\relu$ effect in the deep learning literature, as it helps overcome saturation that the hyperbolic tangent activation functions encounter if the input is very large or very small.

\tableref{tab:result_type_new} also compares the mean accuracy of our \gls{qcmlb} and the \gls{gmlb} method~\cite{minh2019vqa} whereby (1) \gls{gmlb} appears to perform the better on small datasets, such as BACH, IDRiD, or Tools, (2) the proposed \gls{qcmlb} model outperforms \gls{gmlb} on five of the  six datasets, and even on the VQA-Med. A possible explanation for this may be related to the design of \gls{gmlb} that aims at reducing network complexity to prevent overfitting on the small datasets. This may, however, lead to underfitting on large ones instead.

\tableref{tab:result_type_new} shows the mean accuracy and standard errors per-type of question computed from the last 21 epochs for each method on each dataset. As expected, all models perform best on the yes/no question type: the lowest is 66.12 \% on the VQA-V2 dataset by \gls{mutan} without attention mechanism, and the highest is 96.68 \% on the Tools dataset by the proposed approach with the $\relu$ activation function and attention scheme. The most striking observation to emerge from \tableref{tab:result_type_new} is that the proposed models with either $\relu$ or tanh activation functions and attention mechanism have a tendency to outperform state-of-the-art methods on most types of questions on all evaluated datasets.

\begin{figure}[!b]
    \def\subfigsize{0.13\textwidth}
    \def\textsize{0.28\textwidth}
    \def\scalefonesize{0.85}
    \def\scalefonesizetitle{1}   
    
    \centering
    
    \hspace{0.4cm}
    \includegraphics[width=\subfigsize]{figures/sup/img2_what_color_is_the_hydrant_black_and_yellow_in.jpg}
    \begin{tikzpicture}
        \node [text width=\textsize] {{\scalefont{\scalefonesize}Q: What color is the hydrant?\\
        A: Black and yellow\\
        \textcolor{wrongColor}{MUTAN + tanh + Att: Red}\\
        \textcolor{wrongColor}{MLB + tanh + Att: Black and red}\\
        \textcolor{wrongColor}{$\text{G-MLB} + \relu$ + Att: Yellow}\\
        \textcolor{rightColor}{$\text{Ours} + \relu$ + Att: Black and yellow}
        }};
    \end{tikzpicture}
    
    \vspace{0.3cm}
    
    \hspace{0.4cm}
    \includegraphics[width=\subfigsize]{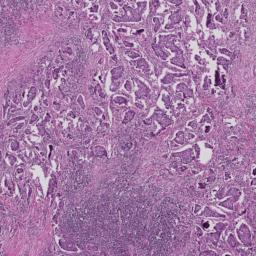}
    \begin{tikzpicture}
        \node [text width=\textsize] {{\scalefont{\scalefonesize}Q: How many classes are there?\\
        A: 2\\
        \textcolor{wrongColor}{MUTAN + tanh + Att: 3}\\
        \textcolor{wrongColor}{MLB + tanh + Att: 3}\\
        \textcolor{wrongColor}{$\text{G-MLB} + \relu$ + Att: 1}\\
        \textcolor{rightColor}{$\text{Ours} + \relu$ + Att: 2}        
        }};
    \end{tikzpicture}

    \caption{Two question-image examples that ultimately fail in current approaches, but succeed in the proposed method. ``+Att'' denotes a method with the attention scheme. $\relu$ and ``tanh'' denote the rectified linear unit and hyperbolic tangent activation functions, respectively. \textcolor{wrongColor}{Red} and \textcolor{rightColor}{Green} denote the wrong and correct predictions, respectively.}
    \label{fig:example_fail}
\end{figure}

\begin{figure*}[!ht]
    \def\subfigsize{0.16\textwidth}
    \def\subfigverlabelsize{0.16\textwidth}
    \def\scalefonesize{0.8}
    \def\scalefonesizetitle{1}

    \centering
    
    \begin{tikzpicture}
        \node [rotate=90, text width=\subfigverlabelsize] {{\scalefont{\scalefonesize}Q: What organ system is imaged?\\A: Skull\\
        \textcolor{rightColor}{No Att.: Skull}\\
        \textcolor{rightColor}{Att.: Skull}}};
    \end{tikzpicture}
    \includegraphics[width=\subfigsize]{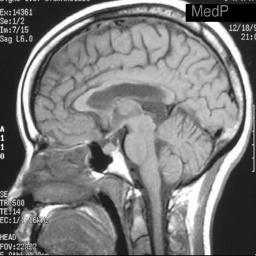}
    \includegraphics[width=\subfigsize]{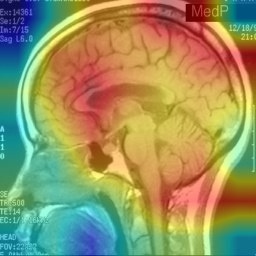}
    \includegraphics[width=\subfigsize]{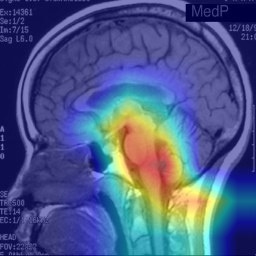}
    \includegraphics[width=\subfigsize]{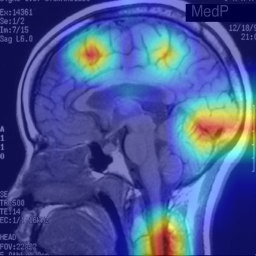}
    \\ 
    
    \begin{tikzpicture}
        \node [rotate=90, text width=\subfigverlabelsize] {{\scalefont{\scalefonesize}Q: Q: What organ system is imaged?\\A: Gastrointestinal\\
        \textcolor{rightColor}{No Att.: Gastrointestinal}\\
        \textcolor{rightColor}{Att.: Gastrointestinal}}};
    \end{tikzpicture}
    \includegraphics[width=\subfigsize]{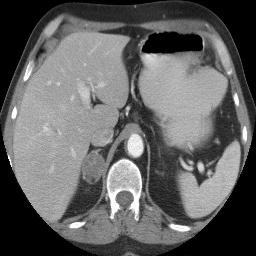}
    \includegraphics[width=\subfigsize]{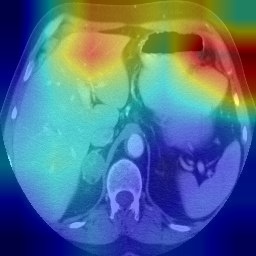}
    \includegraphics[width=\subfigsize]{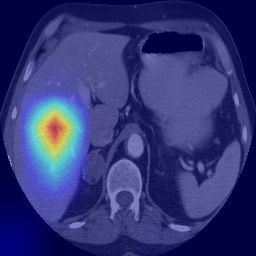}
    \includegraphics[width=\subfigsize]{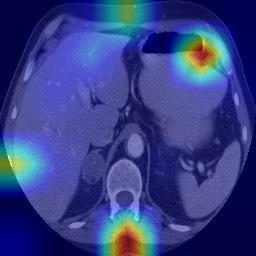}
    \\    
    
    \begin{tikzpicture}
        \node [rotate=90, text width=\subfigverlabelsize] {{\scalefont{\scalefonesize}Q: What color is the hydrant?\\A: Black and yellow\\
        \textcolor{wrongColor}{No Att.: Red}\\
        \textcolor{rightColor}{Att.: Black and yellow}}};
    \end{tikzpicture}    
    \includegraphics[width=\subfigsize]{figures/sup/img2_what_color_is_the_hydrant_black_and_yellow_in.jpg}
    \includegraphics[width=\subfigsize]{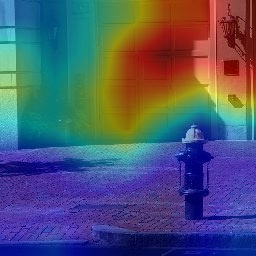}
    \includegraphics[width=\subfigsize]{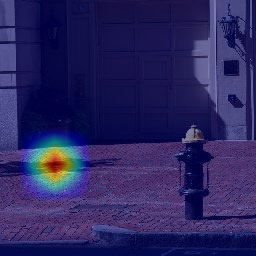}
    \includegraphics[width=\subfigsize]{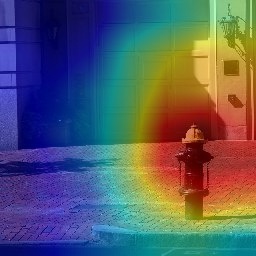}
    \\ 
    
    \begin{tikzpicture}
        \node [rotate=90, text width=\subfigverlabelsize] {{\scalefont{\scalefonesize}Q: What color is the hydrant?\\A: Red\\
        \textcolor{wrongColor}{No Att.: Blue}\\
        \textcolor{wrongColor}{Att.: Red and blue}}};
    \end{tikzpicture}
    \includegraphics[width=\subfigsize]{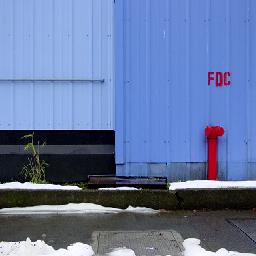}
    \includegraphics[width=\subfigsize]{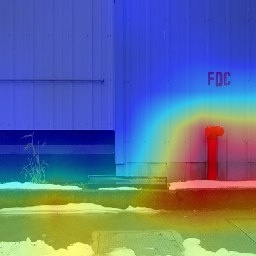}
    \includegraphics[width=\subfigsize]{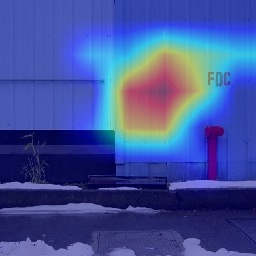}
    \includegraphics[width=\subfigsize]{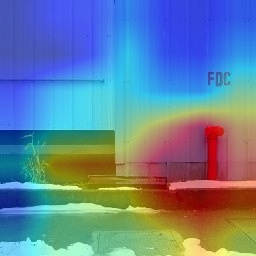}
    \\    

    \begin{tikzpicture}
        \node [rotate=90, text width=0.0] {~\\~\\~\\~};
    \end{tikzpicture}
    \captionsetup[subfigure]{labelformat=empty, justification=centering}
    \begin{subfigure}{\subfigsize}
        \caption{{\scalefont{\scalefonesizetitle} Input image\\~\\~}}  
    \end{subfigure} 
    \begin{subfigure}{\subfigsize}
        \caption{{\scalefont{\scalefonesizetitle}\gls{gradcam} of\\image model\\~}}  
    \end{subfigure} 
    \begin{subfigure}{\subfigsize}
        \caption{{\scalefont{\scalefonesizetitle}\gls{gradcam} of\\\gls{qcmlb}\\(no attention)}}
    \end{subfigure}    
    \begin{subfigure}{\subfigsize}
        \caption{{\scalefont{\scalefonesizetitle}\gls{gradcam} of\\\gls{qcmlb}\\(attention)}}
    \end{subfigure}        
    \caption{Example of medical and natural images and \gls{gradcam} maps from different models on the VQA-Med and VQA datasets. The vertical text on the left shows the pairs of \gls{qa} ground truths used in each row and the predictions of \gls{qcmlb} models without (No Att.) and with attention (Att.) mechanism, respectively. Note that in the last row, although the \gls{qcmlb} model with attention mechanism successfully highlighted the hydrant region, it failed to answer the question correctly (the answer is ``Red'' while the system's answer was ``Red and blue''). \textcolor{wrongColor}{Red} and \textcolor{rightColor}{Green} denote the wrong and correct predictions, respectively.}
    \label{fig:gradcam_natural}
\end{figure*}

\tableref{tab:result_prec_recall} presents the mean precision and recall of all possible answers for each method on each dataset. Here we see that BACH, IDRiD, and Tools show significantly higher precision scores compared to the VQA-Med and two natural VQA datasets across all models. In terms of recall score, only for the IDRiD dataset, are the figures larger than 50 units; while the figures are even smaller than 7 units in BACH and both natural VQA datasets. An explanation as to why the precision and recall scores are low in the VQA-Med, VQA-V1 and VQA-V2 datasets could be that: the number of possible answers in these datasets is very large (greater than 1700), making most models fail to answer all questions types.

From \tableref{tab:result_prec_recall} we also see that most models without attention mechanism tend to have high precision and low recall scores. While the models with attention mechanism seem to perform slightly worse in terms of precision, they outperform by large margins with respect to recall scores models without attention mechanisms. Interestingly, the precision and recall figures of the ``$\text{G-MLB} + \relu$ + Att'' model are greatest and lowest among all models, respectively, in both VQA-V1 and VQA-V2 datasets. A possible explanation for this behavior may be due to the small number of learning parameters in that model that might underfit on large datasets. It is also apparent that the proposed models with attention mechanism perform the best in most tasks in both evaluated metrics.

\subsection{Qualitative Results} 

\begin{figure*}[!ht]
    \def\subfigsize{0.16\textwidth}
    \def\subfigverlabelsize{0.16\textwidth}
    \def\scalefonesize{0.8}
    \def\scalefonesizetitle{1}

    \centering
    \begin{tikzpicture}
        \node [rotate=90, text width=\subfigverlabelsize] {{\scalefont{\scalefonesize}Q: Are there soft exudates?\\A: No\\
        \textcolor{rightColor}{No Att.: No}\\
        \textcolor{rightColor}{Att.: No}}};
    \end{tikzpicture}
    \includegraphics[width=\subfigsize]{figures/gradcam/IDRiD_01_in.jpg}
    \includegraphics[width=\subfigsize]{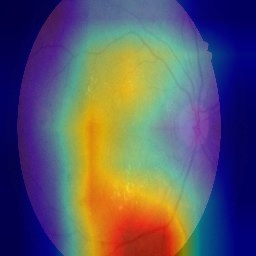}
    \includegraphics[width=\subfigsize]{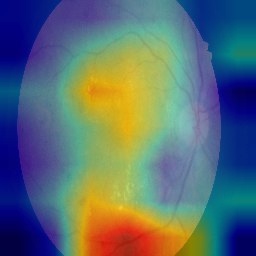}
    \includegraphics[width=\subfigsize]{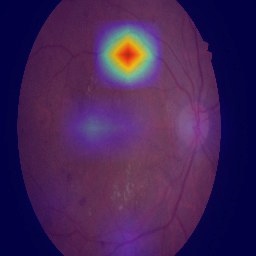}
    \\

    \begin{tikzpicture}
        \node [rotate=90, text width=\subfigverlabelsize] {{\scalefont{\scalefonesize}Q: Are there haemorrhages?\\A: Yes\\
        \textcolor{rightColor}{No Att.: Yes}\\
        \textcolor{rightColor}{Att.: Yes}}};
    \end{tikzpicture}
    \includegraphics[width=\subfigsize]{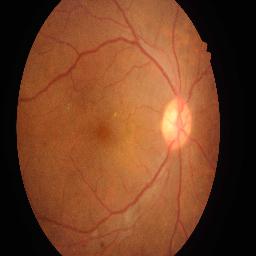}
    \includegraphics[width=\subfigsize]{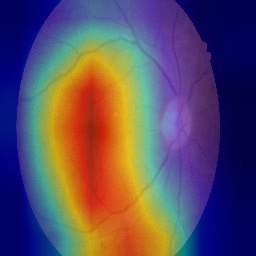}
    \includegraphics[width=\subfigsize]{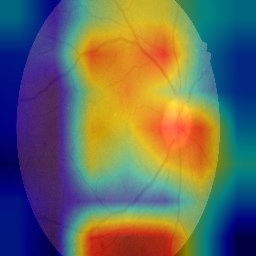}
    \includegraphics[width=\subfigsize]{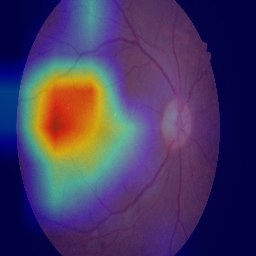}
    \\

    \begin{tikzpicture}
        \node [rotate=90, text width=\subfigverlabelsize] {{\scalefont{\scalefonesize}Q: How many classes are there?\\A: 2\\
        \textcolor{wrongColor}{No Att.: 1}\\
        \textcolor{rightColor}{Att.: 2}}};
    \end{tikzpicture}
    \includegraphics[width=\subfigsize]{figures/gradcam/A05_idx-35648-23728_ps-4096-4096_in.jpg}
    \includegraphics[width=\subfigsize]{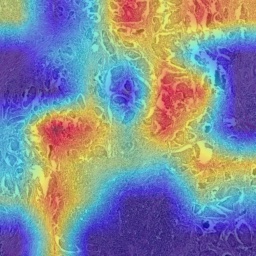}
    \includegraphics[width=\subfigsize]{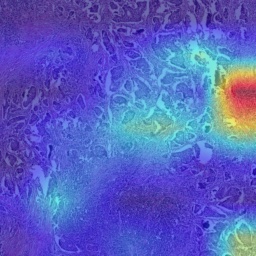}
        \includegraphics[width=\subfigsize]{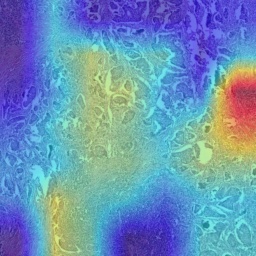}
    \\

    \begin{tikzpicture}
        \node [rotate=90, text width=\subfigverlabelsize] {{\scalefont{\scalefonesize}Q: Is there more of normal than benign?\\A: Yes\\
        \textcolor{rightColor}{No Att.: Yes}\\
        \textcolor{rightColor}{Att.: Yes}}};
    \end{tikzpicture}
    \includegraphics[width=\subfigsize]{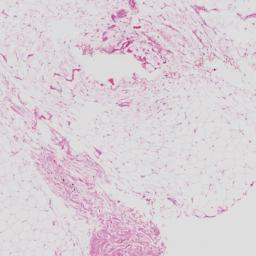}
    \includegraphics[width=\subfigsize]{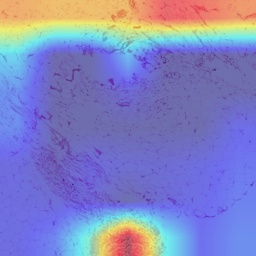}
    \includegraphics[width=\subfigsize]{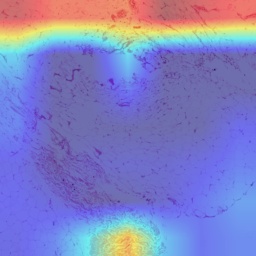}
    \includegraphics[width=\subfigsize]{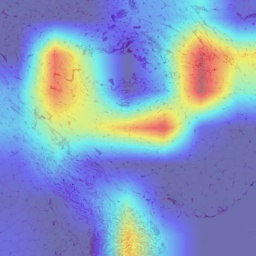}
    \\

    \begin{tikzpicture}
        \node [rotate=90, text width=\subfigverlabelsize] {{\scalefont{\scalefonesize}Q: Which tool has pointed tip on the left?\\A: \textit{n.a.}\\
        \textcolor{wrongColor}{No Att.: Hook}\\
        \textcolor{wrongColor}{Att.: Hook}}};        
    \end{tikzpicture}
    \includegraphics[width=\subfigsize]{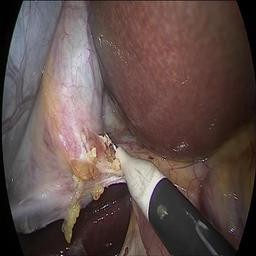}
    \includegraphics[width=\subfigsize]{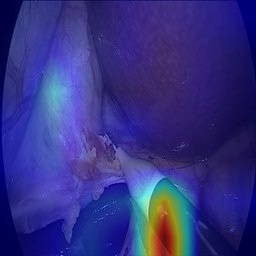}
    \includegraphics[width=\subfigsize]{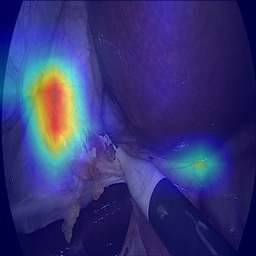}
    \includegraphics[width=\subfigsize]{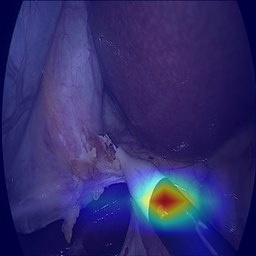}
    \\

    \begin{tikzpicture}
        \node [rotate=90, text width=\subfigverlabelsize] {{\scalefont{\scalefonesize}Q: Is grasper in ``(0,~0) to (32,~32)''?\\
        A: No\\
        \textcolor{rightColor}{No Att.: No}\\
        \textcolor{rightColor}{Att.: No}}};        
    \end{tikzpicture}
    \includegraphics[width=\subfigsize]{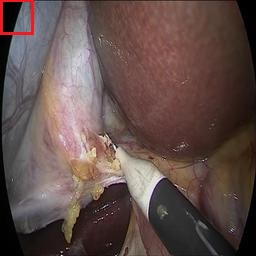}
    \includegraphics[width=\subfigsize]{figures/gradcam/v05_011950_cnn.jpg}
    \includegraphics[width=\subfigsize]{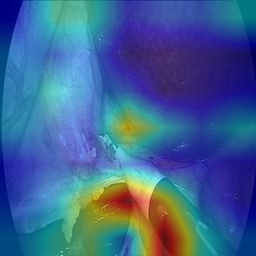}
    \includegraphics[width=\subfigsize]{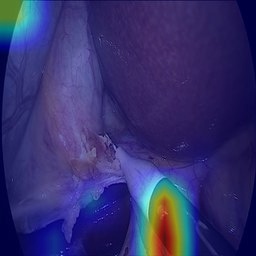}
    \\

    \begin{tikzpicture}
        \node [rotate=90, text width=0.0] {~\\~\\~\\~};
    \end{tikzpicture}
    \captionsetup[subfigure]{labelformat=empty, justification=centering}
    \begin{subfigure}{\subfigsize}
        \caption{{\scalefont{\scalefonesizetitle} Input image\\~\\~}}  
    \end{subfigure} 
    \begin{subfigure}{\subfigsize}
        \caption{{\scalefont{\scalefonesizetitle}\gls{gradcam} of\\image model\\~}}  
    \end{subfigure} 
    \begin{subfigure}{\subfigsize}
        \caption{{\scalefont{\scalefonesizetitle}\gls{gradcam} of\\\gls{qcmlb}\\(no attention)}}
    \end{subfigure}    
    \begin{subfigure}{\subfigsize}
        \caption{{\scalefont{\scalefonesizetitle}\gls{gradcam} of\\\gls{qcmlb}\\(attention)}}
    \end{subfigure}        
    \caption{Example images and \gls{gradcam} maps from different models  on the IDRiD, BACH and Tools datasets. The vertical text on the left shows the pairs of \gls{qa} ground truths used in each row and the predictions of \gls{qcmlb} models without (No Att.) and with attention (Att.) mechanism, respectively. Notice that in the last row, \gls{qcmlb} with attention mechanism put a focus on the top left corner where a location question about ``(0,~0) to (32,~32)'' is asked. Here, (0, 0) and (32, 32) denote the $(x, y)$ coordinates of the corners of the region of interest. Note that in the penultimate row, both networks (with and without the attention mechanism) failed to answer the question (the answer is ``\textit{n.a.}'' while the systems' answers were ``Hook''). \textcolor{wrongColor}{Red} and \textcolor{rightColor}{Green} denote the wrong and correct predictions, respectively.}
    \label{fig:gradcam}
\end{figure*}

In \figref{fig:example_fail} we highlight two question-image examples where existing methods fail to yield correct answers, but where our proposed approach does. Specifically, with the questions ``What color is the hydrant?'' and ``How many classes are there?'', the proposed model answers correctly with the corresponding predictions ``Black and yellow'' and ``2'', while the other methods fail to answer correctly. In the first example, both answers made by ``MUTAN + tanh + Att'' and ``MLB + tanh + Att'' models include ``red'' which is the ground color, while ``yellow'' -- hydrant cover color -- is predicted by the ``G-MLB + RELU + Att''. In the second example, current methods fail again by predicting ``3'', ``3'' and ``1''; however the correct answer is ``2''. It is important to emphasize that these examples are randomly selected from the test set and may not be representative of the overall answering performance. However, they provide some insight into how the proposed model works in relation to the current methods.

\figref{fig:gradcam_natural} and \figref{fig:gradcam} depict two and eight examples of natural and medical images and questions, respectively. The input images are found in the first column, and the other columns contain \gls{gradcam} maps illustrating the models' focus. Column two contains \gls{gradcam} maps computed from the image model, columns three and four contain \gls{gradcam} maps computed from the \gls{qcmlb} model with and without the attention mechanism, respectively.

The first and second rows in \figref{fig:gradcam_natural} illustrate the \gls{gradcam} maps of the image model and the proposed \gls{qcmlb} without and with attention mechanism for the question ``What organ system is imaged.'' It can be seen in the \figref{fig:gradcam_natural} that image model (pretrained ResNet-152) tends to spread the emphasis over the image, while the \gls{qcmlb} without attention mechanism is prone to put a focus on a single region. What is interesting in the first and second rows is that the \gls{qcmlb} with attention mechanism focus on multiple (four) different regions that are equivalent to the number of glimpses. This may explain why models with attention schemes perform better than models without.

It is apparent from the third and last rows in \figref{fig:gradcam_natural} that to answer the question ``What color is the hydrant?'', the proposed model's focus on the hydrant improved when using the attention scheme (from the second column to the last column). This is especially so in the first row, where the image model pays little attention to the hydrant at all. In both cases, the proposed model without attention mechanism did not highlight the hydrant, though the image model succeeded to do so (see the last row).

It appears in \figref{fig:gradcam} that the models with the attention mechanism produce more focused localization maps as compared to those without attention. Further, when using the attention scheme, the relevant areas in the input images appear to be highlighted to a greater extent. For example, in the last row of \figref{fig:gradcam}, with the question ``Is there a grasper in (0, 0) to (32, 32)?'', we can see that the focus is in the top-left corner, and also on the surgical instrument itself. Both highlights are reasonable and appear to be required in order to answer the given question. 

The connection between the \gls{gradcam} maps and the existence of the object in the highlighted region does not mean that the predictions of the model are always accurate, however. For example, the last row in \figref{fig:gradcam_natural} and the penultimate row in \figref{fig:gradcam} demonstrate cases that the proposed models failed to answer, although succeeded to put a focus on important regions in the corresponding images to answer the questions. This limitation could be further explored in future works.





\section{Conclusion}
\label{sec:conclusion}

We have introduced a novel \textit{question-centric} fusion model
for the \gls{vqa} task. The proposed approach was shown to perform better than existing methods, in almost all cases, on three medical imaging datasets, and two natural image datasets. 

One of the limitations of the VQA Pair Generation methods is the potential of propagating annotation errors on to the VQA models. That is, our model may be trained faulty ground truth if the ground truth annotations from the original datasets have errors too, rendering the QA pairs invalid. As a result, these VQA models, which were trained on faulty annotations, would perform poorly because they suffer from the incorrectness of the training input. A potential solution would be to use strategies as shown in~\cite{Nguyen2019} to reduce the need for large amounts of annotations in VQA models. 

Future work will include understanding the emphasizing function and its regularization properties. A natural progression of this work would also be to \textit{learn} the emphasizing function instead of fixing it though a grid-search procedure. Also, it would be interesting to broaden the present work in various fields with strenuous reasoning queries. Similarly, in our work, questions were generated based on a fixed size window (\textit{i.e.} $32 \times 32$). It would appear natural to want to extend that type of question to arbitrary sizes and that questions be user-defined through a GUI (\textit{e.g.}~users can draw an arbitrary shape on an image). We leave such developments to future efforts.

\bibliographystyle{IEEEtran}
\bibliography{bib}

\end{document}